\documentclass{ieeeaccess}

\usepackage{cite}
\usepackage{amsmath,amssymb,amsfonts}
\usepackage{graphicx}
\usepackage{textcomp}
\usepackage{bm}
\usepackage{url} % Needed for the URL in the manuscript

\def\BibTeX{{\rm B\kern-.05em{\sc i\kern-.025em b}\kern-.08em
T\kern-.1667em\lower.7ex\hbox{E}\kern-.125emX}}

\makeatletter
\def\appendices{\par%
    \def\theHsection{Appendix.\Alph{section}}%
    \xdef\Hy@chapapp{appendix}%
    \setcounter{section}{-1}%
    \setcounter{subsection}{0}%
    \setcounter{subsubsection}{0}%
    \setcounter{paragraph}{0}%
    \def\thesection{\Alph{section}}%
    \def\thesectiondis{APPENDIX \Alph{section}}%
    \refstepcounter{section}%
    \setcounter{section}{0}%
}
\makeatother

\newif\ifarxiv
\arxivtrue   % arXiv version
\begin{document}

\ifarxiv
  \history{}
  \doi{}
\else
  \history{Date of publication xxxx 00, 0000,
           date of current version xxxx 00, 0000.}
  \doi{10.1109/ACCESS.XXXX.XXXXXXX}
\fi

\title{Does 3D Pose Improve Action Valuation in Football? An Empirical Study of Pose Representations and Learning Objectives}

\author{
\uppercase{Rikuhei Umemoto}\authorrefmark{1},
\uppercase{Kenjiro Ide}\authorrefmark{1},
\uppercase{Soham Majumder}\authorrefmark{2},
\uppercase{Ulf Brefeld}\authorrefmark{2}, and
\uppercase{Keisuke Fujii}\authorrefmark{1,3}\IEEEmembership{Member, IEEE}
}

\address[1]{Graduate School of Informatics, Nagoya University, Furo-cho, Chikusa, Nagoya, Aichi, Japan}
\address[2]{Institute of Computer Science, Leuphana University of L\"uneburg, L\"uneburg, Germany}
\address[3]{Information Technology Center, Nagoya University, Furo-cho, Chikusa, Nagoya, Aichi, Japan}

\tfootnote{This work was financially supported by JST SPRING Grant Number JPMJSP2125 and JSPS KAKENHI Grant Number 23H03282. R. Umemoto would like to take this opportunity to thank the “THERS Make New Standards Program for the Next Generation Researchers”.}

\markboth
{Umemoto \headeretal: 3D Pose and Action Valuation in Football}
{Umemoto \headeretal: 3D Pose and Action Valuation in Football}

\corresp{Corresponding author: Keisuke Fujii (email: fujii@i.nagoya-u.ac.jp)}

\begin{abstract}
Football action valuation estimates players' contributions to future scoring. Existing multi-agent deep reinforcement learning (MADRL) models use 2D positions and velocities, omitting posture information relevant to imminent actions. We investigate whether adding 3D pose to a SARSA-based model improves observed-action values relative to attack returns: goals, conceded goals or final expected possession value. Using UEFA Champions League 2022/23 event, tracking and 3D pose data, we compare six representations (no pose, raw joints, torso and head orientation, and spatial-temporal graph convolutional network (ST-GCN) embeddings) across five learning settings and three seeds. The settings vary temporal-difference targets and action-prediction loss, with Monte Carlo return regression as a reference. Across 50 on-ball and off-ball comparisons, pose robustly improved return prediction in only three, all using torso and head orientation, while worsening it in twelve, mostly with ST-GCN. Training choices had greater effects: removing the action-prediction loss raised on-ball skill of position-only values from 0.003 to 0.040 and their correlation with expected goals from 0.23 to 0.60; the improvement in return prediction persisted at matched training length. Separately, pose improved prediction of the on-ball action 0.2 s ahead (macro AUC 0.914 to 0.934 with raw joints), while the passer's pose during execution improved pass-success prediction (AUC 0.844 to 0.856). Within this training protocol, pose informed imminent actions but rarely improved observed-action values.
\end{abstract}

\begin{keywords}
action valuation, deep reinforcement learning, multi-agent, 3D pose, graph convolutional network, sports analytics, football
\end{keywords}

\titlepgskip=-15pt
\maketitle

\ifarxiv
\begin{center}
\footnotesize
\textit{
This work has been submitted to the IEEE for possible
publication. Copyright may be transferred without notice,
after which this version may no longer be accessible.
}
\end{center}
\fi

% ======================================================================
\section{Introduction}
\label{sec:introduction}

Football analytics increasingly relies on event and tracking data to model player behaviour and team tactics. Event data describe discrete on-ball actions such as passes, shots and defensive events, while tracking data provide the continuous spatial configuration of the players and the ball. Together, these data have enabled models that estimate the value of actions \cite{decroos2019actions,toda2022evaluation,umemoto2022location}, possessions \cite{simpson2022seq2event,yeung2024unveiling} and tactical patterns \cite{anzer2022detection,rahimian2022inferring}.

Action valuation estimates the contribution of each action to future scoring opportunities. Value-based frameworks include expected possession value \cite{fernandez2019decomposing,rahimian2023towards} and reinforcement-learning-based action valuation \cite{liu2020deep,nakahara2023action,rahimian2022beyond,ide2026strategic}. In particular, MADRL formulates football as a sequential decision-making problem with many interacting players and values both on-ball and off-ball actions \cite{nakahara2023action}.

These models describe the state by the positions and velocities of the players and the ball. Position and velocity tell where a player is and how the player moves, but not how the player is prepared to act. Two players at similar locations and speeds may differ in body orientation, balance and readiness: one may face the goal and be ready to receive or shoot, while the other may be turned away from the ball. Such differences can change both the next action and its value, so an evaluation that ignores them may assign the same value to situations that are not equally promising.

Recent progress in player pose estimation makes 3D body-joint information available in football \cite{bekkers2026wide,pinheiro2022body,wear2022learning,xu2025skor,yeung2024autosoccerpose,liu2026poseaware}, but its use in value-based decision models remains limited. It is also unclear how pose should be represented: raw joint coordinates keep the full posture but enlarge the state, hand-crafted orientation features are compact and interpretable but discard most of the posture, and learned embeddings such as ST-GCN \cite{yan2018spatial} compress short movements but may lose information relevant to valuation, especially when the downstream reward is sparse.

This study asks whether adding 3D pose to an existing MADRL action-valuation model \cite{nakahara2023action} makes the values of the observed actions more valid. We keep the action space, reward and sequence model of the prior model and change only the state representation, and we measure validity by comparing the values of the observed actions with the realised returns of the attacks (a goal, a conceded goal or, otherwise, the expected possession value at the end of the attack), separately for on-ball and off-ball time steps, and with an expected-goals model for shots. A negative answer is hard to interpret on its own: pose may carry no useful information, or the information may not reach the values because of how they are learned. We therefore repeat the comparison under five learning settings, and we test separately, without the value model, whether the same pose features carry information about imminent actions and the outcome of a pass. We found that pose robustly improved the values of the observed actions in only three of 50 comparisons, all with torso and head orientation and in two of the five settings, that the learning setting of the prior model limited the values far more than the state did, and that pose did improve the prediction of the on-ball action 0.2~s ahead and the classification of pass success from the pose at the pass.

The contributions of this paper are as follows.
\begin{itemize}
    \item We extend a SARSA-based MADRL action-valuation model \cite{nakahara2023action} to 3D pose and compare raw joints, torso and head orientation \cite{bekkers2026wide}, and ST-GCN embeddings \cite{yan2018spatial} under the same protocol, evaluating the values of the observed actions against the realised returns and expected goals with match-level bootstrap confidence intervals and the agreement of three seeds.
    \item We show, across five learning settings and 50 comparisons, that pose robustly brings the action values closer to the realised returns in only three comparisons, all with torso and head orientation and none under Monte Carlo return regression \cite{sutton2018reinforcement}, and that ST-GCN embeddings often make them worse.
    \item We show that the action-prediction loss of the prior model lowers the validity of the values of the observed actions, also at matched training length, and that the target of its public implementation can prevent learning.
    \item We show that the same pose features improve the prediction of the on-ball action 0.2~s ahead and that the pose of the passer at the pass is associated with its success, so pose is informative about imminent actions and their execution although, in our data and settings, it rarely improves the values of the observed actions.
\end{itemize}

% ======================================================================
\section{Related Work}
\label{sec:related}

\subsection{Action Valuation in Football}
Action valuation assigns to each action its contribution to future scoring. Event-based frameworks estimate how an action changes the probabilities of scoring and conceding in the near future, as in VAEP \cite{decroos2019actions}, and expected possession value (EPV) models estimate the expected outcome of a possession from the full spatio-temporal configuration of the 22 players and the ball \cite{fernandez2019decomposing,fernandez2021framework}. Reinforcement learning formulates the same problem as the estimation of action values in a sequential decision process: deep reinforcement learning was first used to value actions in ice hockey and football \cite{liu2018deep,liu2020deep}, and multi-agent deep reinforcement learning (MADRL) values both on-ball and off-ball actions of all attacking players \cite{nakahara2023action}. Subsequent studies extended this line to richer decision-making states \cite{ide2026expandable}, to interpretable valuation of all players with strategic multi-agent learning \cite{ide2026strategic}, and to the optimisation of decisions and tactics, such as player decisions in general \cite{rahimian2022beyond}, attacking strategies against low-block defences with offline multi-agent reinforcement learning \cite{pan2026offline}, and corner kicks with graph reinforcement learning \cite{groom2026maximising}. These studies differ in the decision problem and in the state, but, to our knowledge, none of them includes the body pose of the players in the state. The present study keeps the MADRL model of \cite{nakahara2023action} and changes only its state representation.
% Removed until bibliographic data exist (none as of 2026-10-16): "..., and the decision to shoot \cite{verronneaushould}."

Action values have no ground truth, so their evaluation is itself a design choice. Comparisons between valuation frameworks have shown that models that look similar can value the same actions differently \cite{vanroy2020valuing}. The prior MADRL model \cite{nakahara2023action} averaged the values of each player over a season and compared the resulting ranking with season goals, expert ratings, off-ball scoring opportunity (OBSO) \cite{spearman2018beyond} and its extension C-OBSO \cite{teranishi2023evaluation}, finding negative or no correlations, which were read as a different view of player contribution. Such comparisons position a new indicator against existing ones, but they cannot show that one model is more valid than another: the reference indicators rest on their own modelling assumptions and measure other quantities, a negative, null or positive correlation can each be explained after the fact, and player-level rankings are confounded by playing position and the number of attacks. The present study therefore evaluates the values against the realised returns of the same time steps and uses an existing indicator, expected goals, only as a reference for shots (Section~\ref{sec:metrics}). Deep reinforcement learning results are also known to vary strongly between random seeds, so that conclusions from a single run or from confidence intervals that ignore this variation can be misleading \cite{henderson2018deep,agarwal2021deep}; we therefore require differences to agree across seeds.

\subsection{Pose in Sports Analytics}
Recent progress in pose estimation has made 3D body-joint information available in football, and pose has been used to analyse particular actions. Body-orientation angles estimated from broadcast video were associated with the strategies of penalty takers and goalkeepers \cite{pinheiro2022body}. 3D body pose extracted from broadcast footage was used to learn the save technique of professional goalkeepers and to build an expected-saves model \cite{wear2022learning}, and 3D posture sequences of shots were collected and embedded for posture analysis \cite{yeung2024autosoccerpose}. 3D player skeletons and body orientation reconstructed from broadcast video were used to estimate expected goals \cite{xu2025skor,gutierrez2026once}, and head and shoulder angles from pose-enhanced tracking data were used to quantify the visual exploration of players \cite{bekkers2026wide}. Pass success has been modelled from the positions and velocities in tracking data \cite{spearman2017physics,power2017not}, and the orientation of the passer and the receivers, combined with their locations, improved the estimation of which passes are feasible \cite{arbues2020using}. These studies target pose recovery or the analysis of single events such as penalties, saves, shots, scanning and passes; they do not integrate pose into a sequential model that values the actions of all players over an attack. Their findings that pose informs the next shot or pass motivate our analysis of imminent outcomes (Section~\ref{sec:nearterm-method}), which we perform with the same features as the action-value model.

% \subsection{Skeleton Representation Learning}
Skeleton-based action recognition represents a sequence of body poses as a graph. ST-GCN \cite{yan2018spatial} treats the joints as the vertices and the bones as the edges, and applies graph convolutions over the joints and temporal convolutions over the frames, so that it learns spatial and temporal patterns of the skeleton jointly. We use ST-GCN to obtain a fixed-length embedding of a five-frame window of the target player's pose. Unlike in action recognition, the embedding is not trained to classify the movement itself; it is trained jointly with the value model, only through the action-value objective.

% ======================================================================
\section{Problem Statement}
\label{sec:problem}

We consider attacking sequences of football matches sampled at 5 Hz. At each time step $t$ of a sequence, a target player of the attacking team takes an observed action $a_t$ from 15 classes: idle and movement in eight directions for off-ball players, shot, dribble, through pass, pass and cross for the on-ball player, and a defensive action. As in the prior model, a step is an on-ball step when its observed action is one of the five on-ball actions, i.e., when the target player plays the ball with one of these actions in that step; the on-ball actions are masked at all other steps and the eight movements and idle at on-ball steps, while the defensive action is never masked. Between two on-ball events, the ball carrier is therefore labelled with a movement or idle, and only about 7\% of the time steps are on-ball steps of some player. The input is the state $s_t$ seen from the target player, which consists of the 2D positions and velocities of the ball and all 22 players and, in the pose-aware conditions, a representation of the target player's 3D pose. The output is the vector of action values $Q(s_t,\cdot)$.

The reward follows the prior model \cite{nakahara2023action}: it is zero except at the final time step $T$ of a sequence, where it is $1$ for a goal, $-1$ if the team concedes in the opponent's next attack, and otherwise the expected score at the end of the sequence \cite{fernandez2019decomposing}. The EPV is read from the publicly available grid of $32\times50$ pitch cells used by the OpenSTARLab implementation \cite{yeung2025openstarlab}\footnote{The grid file of the OpenSTARLab implementation is identical to the EPV grid published at \url{https://github.com/Friends-of-Tracking-Data-FoTD/LaurieOnTracking/blob/master/EPV_grid.csv}.} at the location of the last event of the sequence of a fixed set of types (passes, crosses including crossed corners and free kicks, shots including direct free kicks, take-ons, interceptions, clearances, fouls, bad touches, and punches and pick-ups by the goalkeeper; the ball position if the event has no location), so it depends only on this location and is not fitted to our data. With $\gamma=1$, the return from step $t$ is $G_t=\sum_{k=t}^{T} r_k = r_T$. We call $G_t$ the realised return; it is an observed goal or conceded goal for the sequences that end with one, and otherwise this location-based estimate, not an observed number of goals.

As in the prior model, each outfield player of the attacking team (the goalkeeper is excluded) is in turn the target player: a sequence gives one row per such player, which consists of the states seen from this player and the actions of this player, and the model values the actions of the target player of the row it is given. The network has no player identifier; the target player is specified by its input slot and the relative-coordinate construction. All target players share the same network, and no joint policy or explicit coordination between players is optimised. We retain the term MADRL to remain consistent with the prior model \cite{nakahara2023action}.

An evaluation is valid when the value of the observed action agrees with what actually happened: $Q(s_t,a_t)$ should be close to $G_t$, and the value of a shot should increase with the probability that the shot is scored. This criterion concerns only the values of the observed actions. The return of an action that was not taken is not observed, so the values of the other available actions, the ranking of the actions and the causal contribution of a player are not validated by it; values learned from fixed data can be far from correct for actions that are rare or absent in the data \cite{fujimoto2019off}. We regard pose as useful when it improves these properties over the same model without pose, trained in the same way. As a separate task, we ask whether pose predicts imminent events from the same state: the on-ball action taken 0.2~s later and the success of a pass. We do not address pose estimation itself, counterfactual optimal actions, or changes of the reward.

% ======================================================================
\section{Method}
\label{sec:method}

\begin{figure*}[t]
    \centering
    \includegraphics[width=\textwidth]{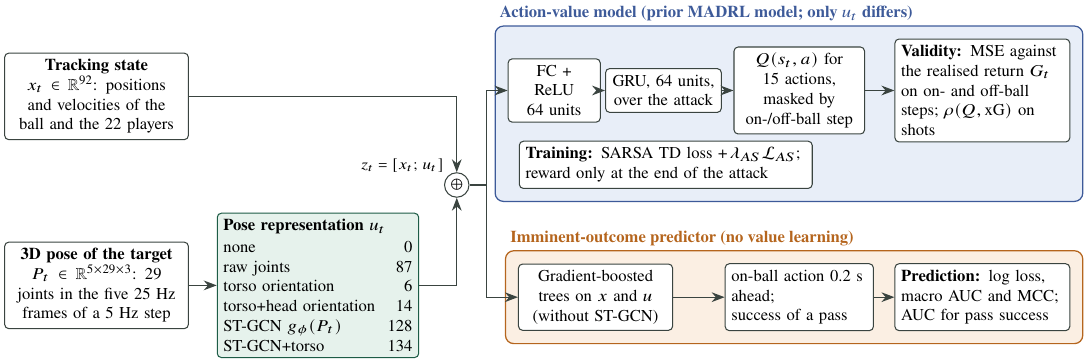}
    \caption{Overview. The 92-dimensional position--velocity state of the ball and all players is concatenated with one of the pose representations of the target player (dimensions on the right): none, raw joints, torso (and head) orientation features, or an ST-GCN embedding with or without torso features. In the action-value model (top), the fused state is projected to 64 dimensions, encoded over the attacking sequence by a GRU, and mapped to action values; unavailable actions are masked according to whether the observed action of the step is an on-ball action. The same state and pose features, except the ST-GCN embeddings, are given to gradient-boosted trees for imminent events (bottom): the on-ball action 0.2~s ahead and, from the pose of the pass step, the success of a pass.}
    \label{fig:overview}
\end{figure*}

\subsection{Overview}
Fig.~\ref{fig:overview} shows the two parts of the study. The action-value model shares the action space, reward and sequence model of the prior model \cite{nakahara2023action}; only the pose part of the state differs between conditions. It is trained under several TD settings (Section~\ref{sec:settings}) and, as a reference, by Monte Carlo return regression \cite{sutton2018reinforcement} (Section~\ref{sec:conditions}). The imminent-outcome predictor uses the same features without any value learning (Section~\ref{sec:nearterm-method}).

\subsection{Action-Value Model}
Let $x_t\in\mathbb{R}^{92}$ be the positions and velocities $(p_x,p_y,v_x,v_y)$ of the ball and the 22 players, in pitch coordinates with the origin at the centre and the attacking team playing towards positive $x$, divided by 52.5 m. The network has no player identifier as input; instead, $x_t$ is built for each target player. Its first 84 dimensions are the other 21 players, with positions relative to the target player and velocities in pitch coordinates, followed by the position and velocity of the ball and, in the last four dimensions, those of the target player in pitch coordinates. The 21 players are ordered by team (attacking team first), then by the role label of the data (DF, FW, GK, MF) and then by the distance to the centre of the goal that the attacking team attacks; the order is computed at every step, so a slot is not tied to one player. The target player is thus defined by the construction of the input rather than by an identity: it is the player in the last four dimensions, the origin of the relative positions of the others and, in the pose-aware conditions, the player whose pose is $u_t$. Even without pose, $x_t$ therefore differs between the target players of the same time step, and the GRU runs over the row of one target player, so its hidden state summarises the history seen from that player. The fused state $z_t$ (Section~\ref{sec:pose-repr}) is projected by a fully connected layer with ReLU to 64 dimensions, encoded by a one-layer GRU \cite{cho2014learning} with 64 hidden units, $h_t=\mathrm{GRU}_\theta(\mathrm{ReLU}(W z_t),h_{t-1})$, and mapped by a fully connected layer to $\mathbf{q}_t=f_\theta(h_t)$, one value per action. As in the implementation of the prior model, this layer has 16 outputs: the 15 actions and a padding class, which is never an observed action and is not used in the evaluation.

The prior model is trained with a SARSA \cite{rummery1994online,sutton2018reinforcement} temporal-difference (TD) loss and an action-prediction loss that uses the values as action scores:
\begin{align}
    y_t &= r_t + \gamma\, m_{t+1}\, Q(s_{t+1},a_{t+1}), \label{eq:td} \\
    \mathcal{L}_{TD} &= \frac{1}{N}\sum_{t} \bigl(y_t - Q(s_t,a_t)\bigr)^2, \\
    \mathcal{L}_{AS} &= -\frac{1}{N}\sum_{t} \log \mathrm{softmax}\bigl(\tilde{\mathbf{q}}_t\bigr)_{a_t}, \\
    \mathcal{L} &= \mathcal{L}_{TD} + \lambda_{AS}\mathcal{L}_{AS},
\end{align}
where $m_{t+1}=1$ if step $t+1$ belongs to the same sequence and $0$ at the end of the sequence, $N$ is the number of real (non-padding) steps, and $\tilde{\mathbf{q}}_t$ is $\mathbf{q}_t$ with the values of the masked actions set to $-99{,}999$, so that the softmax runs over the available actions, the defensive action and the padding output, as in the implementation. We set $\gamma=1$. The target $y_t$ is computed from the values of the same network, without a separate target network, and is treated as a constant in the gradient computation. The prior model also has an L1 penalty on the weights; we set its weight to zero in all settings, because with a positive weight the action prediction was not learned.

\subsection{Learning Settings}
\label{sec:settings}
We train every pose condition under the following settings, which differ in the TD target and the weight $\lambda_{AS}$ of the action supervision loss \cite{nakahara2023action,fujii2023adaptive}.
\begin{itemize}
    \item \textbf{Terminal TD}: \eqref{eq:td} with the standard terminal treatment ($m_{T+1}=0$) over all real steps. With $\lambda_{AS}=0.01$, the weight of the prior model, this is the setting closest to the prior model; we also train it with $\lambda_{AS}=0$.
    \item \textbf{Package TD}: the target as computed by the public implementation of the prior model, over the padded length $t=0,\ldots,T_{\max}-2$, so that the final reward of the longest sequence in a batch is not used and shorter sequences bootstrap from padded steps. We train it with $\lambda_{AS}=0.01$ and $0$. This setting follows the public implementation distributed through OpenSTARLab \cite{yeung2025openstarlab}.
\end{itemize}
With $\lambda_{AS}=0$, the model is trained without an auxiliary action-prediction objective. In all settings, our evaluation is restricted to observed-action values and does not validate the ranking of alternative actions (Fig.~\ref{fig:cases}).

\subsection{Pose Representations}
\label{sec:pose-repr}
For each 5 Hz step, the target player's 29 joints in the five 25 Hz frames of the step are expressed relative to the mid-hip joint and divided by 2.44 m, giving $P_t\in\mathbb{R}^{5\times29\times3}$. The fused state is $z_t=[x_t;\,u_t]$ with one of the following $u_t$.
\begin{itemize}
    \item \textbf{None}: $u_t$ is empty ($z_t\in\mathbb{R}^{92}$).
    \item \textbf{Raw}: the joint coordinates averaged over the valid frames of the step, $u_t\in\mathbb{R}^{87}$.
    \item \textbf{Torso}: the horizontal torso orientation, the unit vector perpendicular to the axis from the left to the right shoulder and pointing to the front of the body, with its dot and cross products with the unit directions to the ball and to the centre of the opponent goal, $u_t\in\mathbb{R}^{6}$.
    \item \textbf{Torso+head}: the torso features plus the horizontal head orientation, perpendicular to the axis from the midpoint of the left eye and ear to that of the right eye and ear, with the sign chosen to agree with the neck-to-nose direction (or with the torso orientation when the nose or neck is missing), and its dot and cross products with the directions to the ball and to the goal and with the torso orientation, $u_t\in\mathbb{R}^{14}$.
    \item \textbf{ST-GCN}: $u_t=g_\phi(P_t)\in\mathbb{R}^{128}$, an ST-GCN embedding.
    \item \textbf{ST-GCN+torso}: $u_t=[g_\phi(P_t);\,u^{\mathrm{torso}}_t]\in\mathbb{R}^{134}$.
\end{itemize}
The orientation features are computed in each frame with the required joints and averaged over the frames of the step. Shoulder and head rotation angles of pose-enhanced tracking data have been used in the same way as the orientation of the body and of the gaze of players \cite{bekkers2026wide}.
The encoder $g_\phi$ applies two ST-GCN blocks \cite{yan2018spatial} on the normalised adjacency matrix of the 29-joint skeleton. Each block consists of a spatial graph convolution followed by a temporal convolution with kernel size 3 and a residual connection; the first block maps the 3 coordinates to 64 channels and the second to 128. The output is averaged over the five frames and the 29 joints. The encoder is initialised randomly and trained jointly with the value model. The handling of missing joints and the details of the encoder are given in Appendices~\ref{app:pose-input} and \ref{app:stgcn}.

\subsection{Imminent-Outcome Prediction}
\label{sec:nearterm-method}
To test whether pose carries information about imminent events independently of value learning, we fit gradient-boosted trees on the same features. For the on-ball action, each on-ball step labelled shot, dribble, through pass, pass or cross is predicted from the state $x_{t-1}$ and pose features $u_{t-1}$ of the previous step of the same sequence, i.e., 0.2~s before the step in which the action is taken; the five frames of step $t-1$ all precede that step. We use the previous step because the pose of the step itself shows the action being executed, which would make the task recognition rather than prediction. For pass success, each on-ball step labelled pass, through pass or cross is matched to the pass event of the attacking team in the time window of the sequence, and the success of that event is predicted from the state and the torso and head features of the passer at that step. Because the pass is played within this step, its pose can include the kick itself, so this analysis relates the outcome of a pass to the posture of its execution rather than predicting the outcome before the pass. The ST-GCN embeddings are learned inside the value model and are not used here.

\subsection{Training}
\label{sec:training}
All action-value models are trained with Adam \cite{kingma2015adam} (learning rate $10^{-4}$, weight decay $10^{-5}$) and batches of 64 sequences, for at most 200 epochs with early stopping on the total validation loss $\mathcal{L}$ (patience 5, absolute minimum improvement $10^{-4}$). As in the public implementation of the prior model, which uses the same patience with a minimum improvement of 0.01, the test uses the weights at the end of training, i.e., those of five epochs after the last improvement by more than the minimum. For TD, selecting instead the epoch with the lowest total validation loss would not select on the value error either, because this loss consists of the TD loss, which does not measure the distance to the true values \cite{fujimoto2022should}, and, with $\lambda_{AS}=0.01$, the action-prediction loss. The gradient-boosted trees \cite{friedman2001greedy} (scikit-learn HistGradientBoostingClassifier \cite{pedregosa2011scikit}, at most 500 iterations, learning rate 0.05, internal early stopping on 15\% of the training data) are fitted on the training matches with five model seeds, and their predicted probabilities are averaged.

The minimum improvement had different effects in the settings. Under terminal TD with $\lambda_{AS}=0.01$, it was about 0.5\% of the total validation loss after the first epoch (0.022), and the runs lasted 47 to 134 epochs; MC regression (Section~\ref{sec:conditions}) ran for 8 to 62 epochs with $\lambda_{AS}=0.01$ and 6 to 12 with $\lambda_{AS}=0$. Without the action loss, however, the validation loss of TD is the TD loss alone, which was $2.8$ to $4.3\times10^{-4}$ after the first epoch and decreased by at most $0.47\times10^{-4}$ over the next five epochs. No later epoch therefore counted as an improvement, and all 36 terminal and package TD runs with $\lambda_{AS}=0$ stopped after $1+5=6$ epochs regardless of their curves. In these runs, the validation loss fell by 2 to 12\% from the first to the third epoch and then varied within 6\%, but it was lowest in the last epoch in 12 of the 36 runs, so a further slow decrease cannot be excluded; the last weights were at most 5.1\% above the lowest validation loss of their run. These curves therefore do not show that the values had converged, and the results of TD without the action loss are those after six epochs. They do show that these values had been learned to some extent: after six epochs, they were closer to the returns than those with the action loss after 47 to 134 epochs (Section~\ref{sec:validity}).

% ======================================================================
\section{Experimental Setup}
\label{sec:setup}

\subsection{Dataset}
We used StatsPerform event data and Hawk-Eye tracking and 3D pose data (29 joints, 25 Hz) of UEFA Champions League 2022/23 matches. Events were synchronised with the tracking data using ELASTIC \cite{kim2025elastic}, and the state--action--reward data followed the OpenSTARLab format \cite{yeung2025openstarlab}. We removed stoppages, intervals affected by substitutions, dismissals or tracking errors, and sequences without usable pose, aligned the attacking direction, and aggregated the data to 5 Hz. Padding players that represent empty roster slots were excluded from event matching and from training. As in the prior model, the action-value model used the attacking sequences in which at least two attackers entered the final third (17.5 m beyond the centre line), about two thirds of the sequences. Each sequence was expanded into one row per attacking player; on-ball steps were counted only in the row of the ball carrier.

The matches were split chronologically by competition phase into training, validation and test sets, with no match in more than one set; a further set of matches reserved for pose-encoder pretraining was not used for the action-value models or the imminent-outcome predictors; only the expected-goals model (Section~\ref{sec:metrics}) was fitted on the shots of all matches other than the test matches, including these. The training, validation and test sets contained 76, 15 and 12 matches with 6,382, 1,202 and 971 attacking sequences and 616,731, 117,097 and 97,100 time steps, respectively; expanded into the rows of the attacking players, they gave 43,812, 8,506 and 6,801 on-ball steps and 6,112,551, 1,160,946 and 962,800 off-ball steps. The training set was used to fit all models, the validation set only for early stopping and for the selection of one hyperparameter of the analysis in Section~\ref{sec:discussion}, and the test set only for the final evaluation.

\subsection{Compared Conditions}
\label{sec:conditions}
We compared six conditions: none (baseline), raw, torso, torso+head, ST-GCN and ST-GCN+torso. Each condition was trained with three random seeds (42, 43, 44) under each learning setting, and each pose condition was compared with the baseline of the same setting.

As a reference outside the proposed model, every condition was also trained by Monte Carlo (MC) return regression (MC regression for short) \cite{sutton2018reinforcement}, in which the bootstrapped target $y_t$ is replaced by the observed return $G_t$ (Section~\ref{sec:problem}), so that $\mathcal{L}_{MC}=\frac{1}{N}\sum_t (G_t-Q(s_t,a_t))^2$ replaces $\mathcal{L}_{TD}$, with $\lambda_{AS}=0.01$ and $0$. MC regression does not propagate values through the model's own estimates and thus serves to test whether the results depend on bootstrapping. Package TD with $\lambda_{AS}=0.01$ did not learn (Section~\ref{sec:validity}), which leaves five settings for the comparison: terminal TD with $\lambda_{AS}=0.01$ and $0$, package TD with $\lambda_{AS}=0$, and MC regression with $\lambda_{AS}=0.01$ and $0$.

\subsection{Evaluation Metrics}
\label{sec:metrics}
The primary metric is the mean squared error between the value of the observed action and the realised return, $\mathrm{MSE}=\frac{1}{N}\sum_t (Q(s_t,a_t)-G_t)^2$, computed separately for on-ball and off-ball steps of the test matches. Because the conditional mean of the return minimises the expected squared error, the difference of the MSE of two models on the same steps estimates the difference of their mean squared distances from the true action values, without access to these values. The TD loss does not have this property, because a small Bellman error does not imply values close to the true ones \cite{fujimoto2022should}, so we do not use it to compare the models. We report it as the skill $1-\mathrm{MSE}/\mathrm{Var}(G)$, which compares the model with a constant prediction of the mean return, and as the difference $\Delta$ of the MSE of the seed-averaged values from that of the baseline on the same steps, with 95\% confidence intervals from 5,000 bootstrap resamples of test matches \cite{efron1993introduction}. These intervals treat the seed-averaged values as fixed and do not include the variation between seeds. We therefore also computed the difference for each seed against the baseline of the same seed, and call a difference robust only if its interval excludes zero and its sign is the same for all three seeds.

We do not use correlations of season-averaged player values with existing indicators such as OBSO \cite{spearman2018beyond} or C-OBSO \cite{teranishi2023evaluation} as evidence of validity, for the reasons given in Section~\ref{sec:related}. As a reference for on-ball values, we compared the value of each observed shot, $Q(s_t,\mathrm{shot})$, with the expected goals (xG) of the same shot. Because the event data contain no xG, we fitted a logistic-regression xG model \cite{eggels2016expected} on the 2,756 shots (own goals excluded) of the 107 non-test matches, with the distance and log-distance to the goal centre, the angle subtended by the posts, and indicators for headers, penalties, direct free kicks, set pieces and fast breaks taken from the event qualifiers. On the 299 test shots, its AUC for goals was 0.75 and its log loss 0.274, against 0.311 for a constant prediction. Each shot step of the action-value model was matched to the shot event of the same match inside the time window of its sequence, by order or, when the counts differed, by location within 10 m; all 250 shot steps were matched to 243 events with 27 goals. We use xG as an auxiliary reference for the scoring potential of observed shots and, with a positive association specified in advance, report the Spearman correlation $\rho(Q,\mathrm{xG})$ with match-bootstrap confidence intervals. xG is not a ground truth for action values and does not use pose.

For imminent outcomes, the on-ball action prediction was evaluated by the log loss and the one-vs-rest macro AUC, and, because passes dominate, also by the macro average precision (PR-AUC), macro-F1 and the Matthews correlation coefficient (MCC) \cite{matthews1975comparison} of the class with the largest ratio of predicted probability to its training share. Pass success was evaluated by the AUC and the log loss. Differences from the state-only model are reported with 95\% confidence intervals from 2,000 bootstrap resamples of test matches, together with the number of the five model seeds for which the difference has the same sign.

For the qualitative examples, the cases were selected by a rule fixed before inspection. We compared the baseline with the pose condition that had the best on-ball primary metric under terminal TD with $\lambda_{AS}=0.01$. For each test step, the values were averaged over the three seeds, and $\Delta_t=|Q_{\mathrm{pose}}(s_t,a_t)-G_t|-|Q_{\mathrm{none}}(s_t,a_t)-G_t|$ was computed. Success cases were the steps with the smallest $\Delta_t$ and failure cases those with the largest $\Delta_t$, taking at most one step per sequence within each of the four lists (success and failure, on-ball and off-ball) and only steps without missing pose. Because a single outcome is noisy, these cases illustrate how pose changes the values; they are not evidence of validity.

\subsection{Implementation Details}
Experiments were run with PyTorch 2.9.1 (CUDA 12.8) and PyTorch Lightning 2.5.6 on NVIDIA A100-SXM4-80GB GPUs, one GPU per run.

% ======================================================================
\section{Results}
\label{sec:results}

\subsection{Validity of Action Values}
\label{sec:validity}
\begin{table*}[t]
\centering
\caption{Validity of the action values on the test matches under five learning settings. Skill: $1-\mathrm{MSE}/\mathrm{Var}(G)$, mean over three seeds; $\Delta$: MSE of the seed-averaged values minus that of the baseline of the same setting on the same steps ($\times10^{-4}$, 95\% CI from 5,000 bootstrap resamples of test matches); $\rho(Q,\mathrm{xG})$: Spearman correlation of the seed-averaged shot values with xG on the 250 test shots [95\% CI]. $\mathrm{Var}(G)=3.507\times10^{-2}$ (on-ball) and $3.319\times10^{-2}$ (off-ball). Negative $\Delta$ means closer to the realised return; bold: best skill and $\rho$ within each setting (ties at the shown precision all in bold); $^{\dagger}$: robust difference (CI excludes zero and the same sign for all three seeds), in either direction. Package TD with $\lambda_{AS}=0.01$ did not learn and is not shown.}
\label{tab:validity}
\setlength{\tabcolsep}{4pt}
\begin{tabular}{lccccc}
\hline
 & \multicolumn{2}{c}{On-ball steps} & \multicolumn{2}{c}{Off-ball steps} & Shots \\
Condition & Skill & $\Delta$ [95\% CI] & Skill & $\Delta$ [95\% CI] & $\rho(Q,\mathrm{xG})$ [95\% CI] \\
\hline
\multicolumn{6}{l}{\textit{Terminal TD, $\lambda_{AS}=0.01$ (prior model)}} \\
none (baseline) & $0.003$ & -- & $-0.057$ & -- & 0.23 [0.15, 0.32] \\
raw & $0.000$ & $+0.40$ [$-1.21$, $+1.94$] & $\mathbf{-0.017}$ & $-3.99$ [$-7.01$, $-0.91$] & $\mathbf{0.26}$ [0.16, 0.36] \\
torso & $0.005$ & $-0.89$ [$-1.98$, $+0.19$] & $-0.085$ & $+5.93$ [$+2.90$, $+8.99$] & 0.25 [0.16, 0.35] \\
torso+head & $\mathbf{0.010}$ & $-2.71^{\dagger}$ [$-4.04$, $-1.49$] & $-0.046$ & $-0.50$ [$-3.00$, $+2.40$] & 0.22 [0.14, 0.30] \\
ST-GCN & $0.005$ & $-1.17$ [$-3.16$, $+0.57$] & $-0.043$ & $+4.55$ [$+1.77$, $+6.98$] & 0.17 [0.09, 0.25] \\
ST-GCN+torso & $-0.007$ & $+2.69^{\dagger}$ [$+1.09$, $+4.41$] & $-0.047$ & $+3.82$ [$+0.65$, $+6.91$] & 0.13 [0.06, 0.22] \\
\hline
\multicolumn{6}{l}{\textit{Terminal TD, $\lambda_{AS}=0$}} \\
none (baseline) & $0.040$ & -- & $\mathbf{0.018}$ & -- & $\mathbf{0.60}$ [0.48, 0.70] \\
raw & $0.038$ & $+0.46$ [$-0.08$, $+0.96$] & $\mathbf{0.018}$ & $-0.27$ [$-0.70$, $+0.17$] & 0.57 [0.44, 0.69] \\
torso & $0.038$ & $+0.24$ [$-0.63$, $+1.09$] & $0.016$ & $+0.42$ [$-0.12$, $+0.97$] & 0.53 [0.39, 0.67] \\
torso+head & $\mathbf{0.042}$ & $-1.08$ [$-2.60$, $+0.59$] & $\mathbf{0.018}$ & $-0.29$ [$-1.05$, $+0.49$] & 0.53 [0.37, 0.67] \\
ST-GCN & $0.021$ & $+5.52^{\dagger}$ [$+3.51$, $+7.93$] & $0.002$ & $+4.42^{\dagger}$ [$+1.88$, $+7.17$] & 0.59 [0.43, 0.72] \\
ST-GCN+torso & $0.028$ & $+2.99^{\dagger}$ [$+1.39$, $+4.68$] & $0.011$ & $+1.64^{\dagger}$ [$+0.04$, $+3.17$] & 0.56 [0.40, 0.71] \\
\hline
\multicolumn{6}{l}{\textit{Package TD, $\lambda_{AS}=0$}} \\
none (baseline) & $0.030$ & -- & $0.010$ & -- & 0.56 [0.44, 0.67] \\
raw & $0.026$ & $+1.06$ [$+0.35$, $+1.65$] & $0.008$ & $+0.54$ [$-0.07$, $+1.07$] & 0.54 [0.40, 0.67] \\
torso & $0.029$ & $+0.02$ [$-0.80$, $+0.73$] & $0.008$ & $+0.58$ [$+0.03$, $+1.09$] & 0.52 [0.37, 0.65] \\
torso+head & $\mathbf{0.036}$ & $-2.23^{\dagger}$ [$-3.75$, $-0.74$] & $\mathbf{0.014}$ & $-1.41^{\dagger}$ [$-2.39$, $-0.41$] & 0.51 [0.36, 0.65] \\
ST-GCN & $0.011$ & $+5.31^{\dagger}$ [$+3.58$, $+7.08$] & $-0.012$ & $+6.17^{\dagger}$ [$+3.73$, $+8.56$] & $\mathbf{0.58}$ [0.43, 0.71] \\
ST-GCN+torso & $0.020$ & $+2.33$ [$+0.90$, $+3.50$] & $0.003$ & $+1.83$ [$+0.34$, $+3.13$] & 0.56 [0.40, 0.71] \\
\hline
\multicolumn{6}{l}{\textit{MC regression, $\lambda_{AS}=0.01$}} \\
none (baseline) & $0.034$ & -- & $\mathbf{0.014}$ & -- & $\mathbf{0.64}$ [0.54, 0.73] \\
raw & $0.035$ & $-0.27$ [$-1.40$, $+0.68$] & $0.011$ & $+0.96$ [$-0.05$, $+1.89$] & 0.57 [0.47, 0.67] \\
torso & $0.031$ & $+1.58^{\dagger}$ [$+0.11$, $+3.07$] & $0.012$ & $+1.08$ [$-0.25$, $+2.44$] & 0.53 [0.42, 0.64] \\
torso+head & $\mathbf{0.038}$ & $-1.32$ [$-3.42$, $+0.82$] & $0.012$ & $+0.52$ [$-1.31$, $+2.51$] & 0.56 [0.45, 0.67] \\
ST-GCN & $0.027$ & $+0.87$ [$-2.63$, $+4.07$] & $-0.045$ & $+13.08^{\dagger}$ [$+9.38$, $+16.59$] & 0.40 [0.31, 0.50] \\
ST-GCN+torso & $0.021$ & $+1.75$ [$-2.95$, $+6.00$] & $-0.058$ & $+16.63^{\dagger}$ [$+11.98$, $+20.84$] & 0.47 [0.38, 0.54] \\
\hline
\multicolumn{6}{l}{\textit{MC regression, $\lambda_{AS}=0$}} \\
none (baseline) & $0.042$ & -- & $0.016$ & -- & $\mathbf{0.61}$ [0.50, 0.70] \\
raw & $0.040$ & $+0.35$ [$-0.34$, $+1.04$] & $0.014$ & $+0.29$ [$-0.54$, $+1.01$] & 0.54 [0.44, 0.65] \\
torso & $0.039$ & $+1.05^{\dagger}$ [$+0.08$, $+2.03$] & $0.013$ & $+1.05^{\dagger}$ [$+0.10$, $+1.93$] & 0.52 [0.40, 0.63] \\
torso+head & $0.042$ & $-0.23$ [$-1.69$, $+1.19$] & $0.016$ & $+0.26$ [$-1.16$, $+1.53$] & 0.47 [0.33, 0.61] \\
ST-GCN & $0.039$ & $+0.91$ [$-0.72$, $+2.63$] & $0.018$ & $-0.51$ [$-1.96$, $+0.90$] & 0.43 [0.28, 0.57] \\
ST-GCN+torso & $\mathbf{0.044}$ & $-0.33$ [$-1.91$, $+1.21$] & $\mathbf{0.019}$ & $-1.00$ [$-2.46$, $+0.40$] & 0.50 [0.38, 0.60] \\
\hline
\end{tabular}
\end{table*}

\begin{figure*}[t]
\centering
\includegraphics[width=\textwidth]{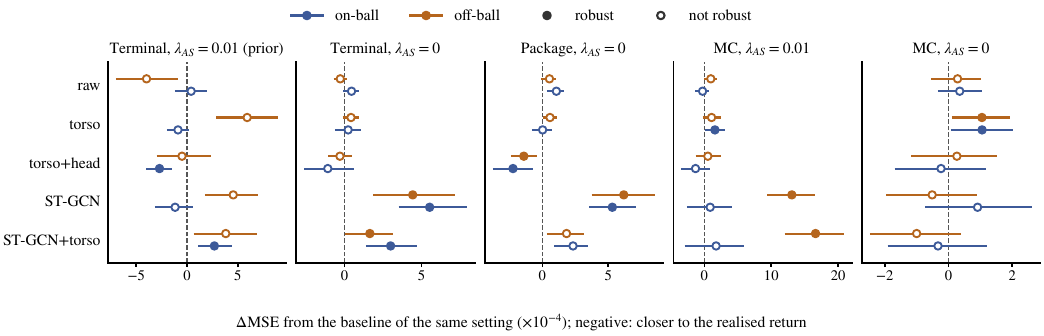}
\caption{Difference in MSE between each pose condition and the baseline of the same setting on the test matches ($\times10^{-4}$, values averaged over three seeds), with 95\% CIs from 5,000 bootstrap resamples of test matches. Filled markers: robust differences (CI excludes zero and the same sign for all three seeds). The horizontal scales differ between panels. Package TD with $\lambda_{AS}=0.01$ did not learn and is not shown.}
\label{fig:forest}
\end{figure*}

Table~\ref{tab:validity} and Fig.~\ref{fig:forest} show the first main result. Across the five settings, five pose conditions and on-ball and off-ball steps, pose brought the values robustly closer to the realised returns in three of the 50 comparisons and made them robustly worse in twelve. All three improvements came from torso+head, on on-ball steps under terminal TD with $\lambda_{AS}=0.01$ ($\Delta=-2.71\times10^{-4}$) and on both kinds of steps under package TD with $\lambda_{AS}=0$ ($-2.23$ and $-1.41\times10^{-4}$). The on-ball difference of torso+head was negative in all five settings, but under terminal TD with $\lambda_{AS}=0$, which differs from package TD only in the treatment of the end of the sequences, its interval included zero ($-1.08$ [$-2.60$, $+0.59$]), and under both MC settings its seeds also disagreed in sign; its off-ball difference was negative in three settings and positive in two. Nine of the twelve robust deteriorations came from the ST-GCN embeddings: ST-GCN on both kinds of steps under terminal and package TD with $\lambda_{AS}=0$ ($+4.4$ to $+6.2\times10^{-4}$), ST-GCN+torso on both kinds of steps under terminal TD with $\lambda_{AS}=0$ ($+3.0$ and $+1.6$) and on on-ball steps under terminal TD with $\lambda_{AS}=0.01$ ($+2.7$), and both on off-ball steps under MC regression with $\lambda_{AS}=0.01$ ($+13.1$ and $+16.6$). The other three came from torso under MC regression ($+1.05$ to $+1.58$), and the raw joints were within the noise of the baseline in every comparison. On the shots, no pose condition had a correlation with xG above the interval of its baseline, and the ST-GCN conditions had the lowest correlations under three settings.

The confidence intervals over matches alone would have suggested more effects. Under terminal TD with $\lambda_{AS}=0.01$, the off-ball intervals of four pose conditions exclude zero (raw $-3.99$, torso $+5.93$, ST-GCN $+4.55$ and ST-GCN+torso $+3.82\times10^{-4}$), but every condition in this setting, including the baseline, had a larger off-ball error than the constant prediction, and the difference was driven by one baseline seed: the off-ball MSE of the baseline was $3.365$, $3.408$ and $3.755\times10^{-2}$ for seeds 42, 43 and 44, and the differences of raw and torso from the baseline seed with the same number changed sign across the seeds ($+3.0$, $-5.7$, $-37.7$ and $+19.5$, $+32.7$, $-24.4\times10^{-4}$). The same seed also reversed the sign of the differences of both ST-GCN conditions. This is why we require the same sign for all seeds.

The learning setting changed the values far more than pose did. Under terminal TD with $\lambda_{AS}=0.01$, the setting closest to the prior model, the baseline was barely better than a constant prediction on on-ball steps (skill 0.003) and worse on off-ball steps ($-0.057$), and its shot values correlated only weakly with xG (0.23). Removing the action-prediction loss raised the skill to 0.040 and 0.018 and the correlation with xG to 0.60; the MSE of the baseline decreased by $12.2\times10^{-4}$ [$8.4$, $16.0$] on on-ball and $12.1\times10^{-4}$ [$8.8$, $15.2$] on off-ball steps, with the same sign for all seeds. MC regression reached a similar level with either weight (on-ball skill 0.034 and 0.042, correlation with xG 0.64 and 0.61). Package TD with $\lambda_{AS}=0.01$ did not learn in any of the 18 runs: from the first epoch, all action values moved together towards large negative values (mean value of the observed action $-7.7$ after the first and $-36$ after the sixth epoch for the baseline, $-11$ to $-26$ at the end for the ST-GCN conditions), and early stopping ended the runs after six to 25 epochs. With $\lambda_{AS}=0$, the same target stayed in the range of the returns within the six epochs trained; like every TD run without the action loss, these runs stopped after six epochs (Section~\ref{sec:training}), so whether their values would drift with longer training is untested. This limits the weight of the two improvements of torso+head in this setting. Its improvement in the prior setting was about a fifth of the effect of removing the action-prediction loss on the baseline ($-2.71$ against $-12.2\times10^{-4}$), and its values there remained less skilful than the position-only values without that loss (0.010 against 0.040). Thus pose improved the values only in a few comparisons of two TD settings and in neither MC setting.

\subsection{Pose and Imminent Outcomes}
\label{sec:nearterm}
\begin{table}[t]
\centering
\caption{Prediction of the on-ball action 0.2~s ahead on the test matches (6,388 steps; pass 87\%, cross 5\%, dribble 4\%, shot 4\%, through pass 0.3\%) by gradient-boosted trees on the state and pose features of the previous step. Differences from the state-only model with 95\% CIs over test matches; in parentheses, the number of the five model seeds with the same sign; bold: best value of each column.}
\label{tab:imminent}
\setlength{\tabcolsep}{2pt}
\begin{tabular}{lccc}
\hline
Features & Log loss & Macro AUC & MCC \\
\hline
state only & 0.304 & 0.914 & 0.424 \\
\hline
+torso & $-0.013$ & $+0.009$ (5/5) & $+0.013$ (5/5) \\
 & [$-0.017$, $-0.009$] & [$+0.006$, $+0.013$] & [$-0.001$, $+0.027$] \\
+torso+head & $-0.013$ & $+0.010$ (5/5) & $+0.014$ (5/5) \\
 & [$-0.016$, $-0.009$] & [$+0.007$, $+0.012$] & [$+0.003$, $+0.024$] \\
+raw & $\mathbf{-0.038}$ & $\mathbf{+0.020}$ (5/5) & $\mathbf{+0.089}$ (5/5) \\
 & [$-0.043$, $-0.032$] & [$+0.011$, $+0.028$] & [$+0.075$, $+0.101$] \\
\hline
\end{tabular}
\end{table}

Table~\ref{tab:imminent} shows the second main result: the same pose features that did not robustly improve the action values improved the prediction of the on-ball action 0.2~s ahead. The raw joints reduced the log loss by $0.038$ and raised the macro AUC from 0.914 to 0.934 and the MCC from 0.424 to 0.512, for all five model seeds; the torso and head orientation features gave smaller improvements, with the same sign of the AUC and MCC differences for all five model seeds and intervals excluding zero except for the MCC of torso. The gains also appeared in the metrics that weight the rare actions: with the raw joints, the macro PR-AUC rose by $+0.062$ [$+0.027$, $+0.085$] and the macro-F1 by $+0.069$ [$+0.054$, $+0.085$], and the PR-AUC of shots rose from 0.66 to 0.78 (base rate 0.04), of dribbles from 0.28 to 0.40 and of crosses from 0.64 to 0.73. Through passes, with 18 test cases, were hardly predicted with any features.

Pose was also associated with the outcome of the pass that it accompanies. For pass success, the matched sequences (1,845 of the 6,332 training and 390 of the 969 test sequences that contain a pass, in which the numbers of passes and pass events agreed) gave 10,034 training and 2,109 test passes, 85\% of them successful. Adding the 14 torso and head features of the passer at the pass step, whose pose can include the kick (Section~\ref{sec:nearterm-method}), to the 92-dimensional state raised the test AUC from 0.844 to 0.856 (difference $+0.012$ [$+0.007$, $+0.018$], higher for each of the five model seeds); the log loss decreased from 0.305 to 0.298 (difference $-0.007$ [$-0.011$, $-0.004$]; constant prediction 0.412).

\subsection{Qualitative Examples}
\begin{figure*}[p]
\centering
\includegraphics[width=0.85\textwidth]{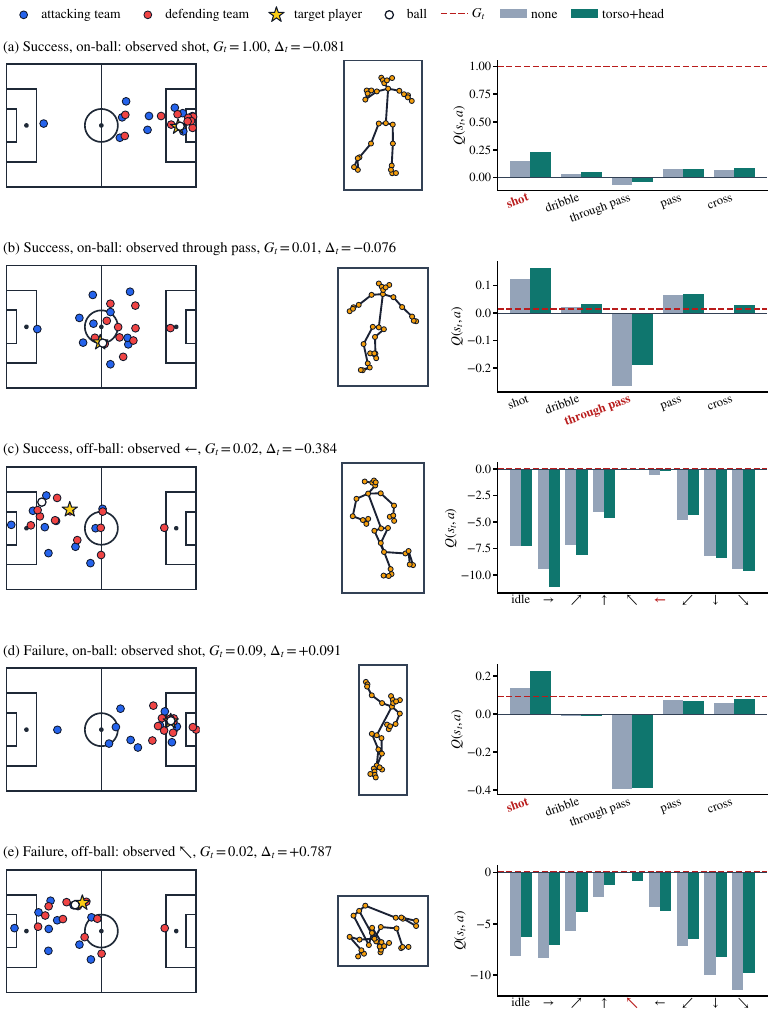}
\caption{Qualitative cases selected by the rule in Section~\ref{sec:metrics} (terminal TD, $\lambda_{AS}=0.01$, torso+head against none, values averaged over three seeds): (a)--(c) success cases with the smallest $\Delta_t$ and (d), (e) failure cases with the largest $\Delta_t$. Left: players and ball, with the attacking team playing from left to right; middle: front view of the target's pose averaged over the frames of the step; right: values of the available actions, with the observed action in red, and the realised return $G_t$. The share of test steps with $\Delta_t<0$ was 0.61 on-ball and 0.47 off-ball.}
\label{fig:cases}
\end{figure*}

Fig.~\ref{fig:cases} shows the cases selected by the rule in Section~\ref{sec:metrics}; torso+head had the best on-ball metric under terminal TD with $\lambda_{AS}=0.01$. In the success cases, torso+head moved the values of the observed actions towards the realised returns: it raised the value of a scored shot from 0.15 to 0.23 and those of a through pass and of an off-ball movement that the baseline valued below the return. In the on-ball failure, it raised the value of an unscored shot in the same way, and in the off-ball failure, the head of the target was below the hips in all frames of the step, a fallen or misestimated pose, and the value of the observed movement fell from 0.04 to $-0.80$. The values of the unobserved off-ball actions, far below the range of the returns for both models, show that with $\lambda_{AS}=0.01$ the action-prediction loss uses the values as action scores (Section~\ref{sec:discussion}).

% ======================================================================
\section{Discussion}
\label{sec:discussion}

\subsection{Why Pose Did Not Improve the Action Values}
Pose carried information about the next action and about the outcome of a pass at its execution (Section~\ref{sec:nearterm}), but not in a form that improved the action values. We discuss two factors that may explain this gap; the experiments support the first, whereas the second is a hypothesis that we did not test directly.

\begin{figure}[t]
\centering
\includegraphics[width=\columnwidth]{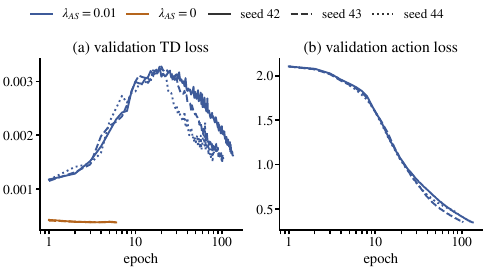}
\caption{Validation losses of the baseline under terminal TD per epoch (logarithmic scale). (a) TD loss with $\lambda_{AS}=0.01$ and $0$; (b) action-prediction loss with $\lambda_{AS}=0.01$. With $\lambda_{AS}=0.01$, early stopping on the total validation loss ended the runs after 83 to 134 epochs, while the TD loss stayed above its first-epoch value; with $\lambda_{AS}=0$, all runs stopped after six epochs, because the minimum improvement of early stopping exceeded the whole decrease of the loss after the first epoch (Section~\ref{sec:training}). The levels of the two TD losses are not comparable, because each is computed from the model's own values.}
\label{fig:curves}
\end{figure}

The first is the learning setting of the prior model. Auxiliary losses can help reinforcement learning when they shape a shared representation \cite{jaderberg2017reinforcement}, but the action-prediction loss uses the values themselves as action scores, so it competes with value learning: under terminal TD with $\lambda_{AS}=0.01$, the weighted gradient of the action loss was 0.43 to 0.96 times that of the TD loss (median over epochs, mean over seeds), against 0.07 to 0.22 under MC regression. Because early stopping monitors the total validation loss, which includes the action loss, training continued for 47 to 134 epochs (Fig.~\ref{fig:curves}), although the validation TD loss rose after the first epoch in every run and, in 14 of the 18 runs, never fell below its first-epoch value (in the other four, it was at most 13\% lower after 52 to 82 epochs). With $\lambda_{AS}=0$, in contrast, every terminal-TD run stopped after six epochs (Section~\ref{sec:training}). To separate the action loss from the training length, we retrained terminal TD with $\lambda_{AS}=0.01$ for exactly six epochs. The baseline then had an on-ball skill of $-0.039$ and an off-ball skill of $-0.249$, below the 0.040 and 0.018 reached without the action loss after the same six epochs; the MSE without the action loss was lower by $24.2\times10^{-4}$ [$19.6$, $28.5$] on on-ball and $82.0\times10^{-4}$ [$76.4$, $87.4$] on off-ball steps, for every seed. The action loss therefore lowered the validity of the values both at matched length and at the length chosen by early stopping.

Training length did change how much information about actions the values hold. We fitted a multinomial logistic regression that predicts the observed on-ball action from the five on-ball action values of the step (a probe \cite{alain2016understanding}), on the 76 training matches, with the inverse regularisation strength chosen from $\{10^{-3},\ldots,10^{2}\}$ by the log loss on the validation matches, and evaluated it once on the test matches (Table~\ref{tab:probe}). After six epochs, the values of the baseline predicted the action with a macro AUC of 0.853; after training to the end, only 0.735. With long training, raw, ST-GCN and ST-GCN+torso had robustly higher AUC than the baseline (0.79 to 0.80), which could be read as pose information reaching the values; at matched length, however, every pose condition was below the baseline, robustly so for raw, ST-GCN and ST-GCN+torso. The apparent advantage of the pose conditions after longer training may partly reflect the decline in the baseline during training, rather than solely information added by pose. A similar pattern was observed with the probe fitted on the validation matches or on a subset of the training matches with a fixed regularisation strength, and with the four on-ball actions other than the rare through pass.

\begin{table}[t]
\centering
\caption{Action information in the values (probe on the five on-ball action values, fitted on the training matches, tested on the 6,801 on-ball test steps): macro AUC and MCC for terminal TD with $\lambda_{AS}=0.01$ trained until early stopping (47--134 epochs) or for six epochs, and with $\lambda_{AS}=0$ (six epochs). Bold: best value of each column; $^{\dagger}$: robust difference from the baseline of the same column, in either direction.}
\label{tab:probe}
\setlength{\tabcolsep}{2.5pt}
\begin{tabular}{lcccccc}
\hline
 & \multicolumn{2}{c}{$\lambda_{AS}=0.01$} & \multicolumn{2}{c}{$\lambda_{AS}=0.01$, 6 ep.} & \multicolumn{2}{c}{$\lambda_{AS}=0$} \\
Condition & AUC & MCC & AUC & MCC & AUC & MCC \\
\hline
none & $0.735$ & $0.131$ & $\mathbf{0.853}$ & $\mathbf{0.270}$ & $0.871$ & $0.266$ \\
raw & $0.792^{\dagger}$ & $0.187$ & $0.818^{\dagger}$ & $0.234^{\dagger}$ & $\mathbf{0.880}$ & $\mathbf{0.296}^{\dagger}$ \\
torso & $0.726$ & $0.108$ & $0.840$ & $0.239^{\dagger}$ & $0.873$ & $0.265$ \\
torso+head & $0.746$ & $0.141$ & $0.830$ & $0.226$ & $0.866$ & $0.283^{\dagger}$ \\
ST-GCN & $0.800^{\dagger}$ & $0.181^{\dagger}$ & $0.800^{\dagger}$ & $0.208^{\dagger}$ & $0.846^{\dagger}$ & $0.258$ \\
ST-GCN+torso & $\mathbf{0.801}^{\dagger}$ & $\mathbf{0.189}$ & $0.794^{\dagger}$ & $0.183^{\dagger}$ & $0.876$ & $0.288$ \\
\hline
\end{tabular}
\end{table}

The second factor may be the distance between the posture and the reward. In the settings without the action loss and with MC regression, the baseline learned the position information well enough for its shot values to agree with xG, and pose robustly improved the values only for torso+head under package TD, whereas the same features improved the prediction of the next action and the classification of pass success at the pass. Body orientation and preparation affect the execution of the next action, but the end of an attack is several actions and several seconds later, and its outcome depends on many later decisions of both teams. A small effect on the next action may therefore be diluted in a sparse and noisy return. Other explanations are not excluded, such as errors and gaps in the estimated poses, the absence of the poses of the other players, the capacity of the model and the amount of data; comparing returns over horizons of different lengths would separate the distance to the reward from these.

\subsection{Learned Pose Embeddings}
The ST-GCN embeddings, trained from random initialisation jointly with the value model, accounted for nine of the twelve robust deteriorations, in four of the five settings, and they lowered the correlation of the shot values with xG under three settings. With $\lambda_{AS}=0.01$, their seed-averaged values of the observed off-ball actions fell outside the range $[-1,1]$ of the returns at 82 and 124 test steps, which no such value of the baseline did. A 128-dimensional embedding trained only through a sparse value signal can fit noise in the training returns: the encoder raised the number of trainable parameters from about 32,000 to 120,000 (Table~\ref{tab:params}).

Two properties of our encoder input may add to this (Appendix~\ref{app:pose-input}). First, the input has no mask: a missing joint is set to the position of the mid-hip, and a missing frame, a step without pose and a padded step are all zeros, so the encoder cannot tell absent pose from a posture with joints at the hip. Second, batch normalisation \cite{ioffe2015batch} inside the encoder computes its statistics over all steps of a batch, including these zero inputs, whose share depends on the lengths of the sequences in the batch, while the evaluation uses the running statistics accumulated during training. Both can make the embedding of the same posture depend on the presence of missing or padded steps rather than on the posture itself, and both affect only the ST-GCN conditions, which used the same joints as raw but, unlike raw, made the values robustly worse. We did not test these explanations; masking the missing joints and padded steps, normalising without them, or replacing batch normalisation by a per-sample normalisation would separate them from the effect of the sparse reward. Constraints on the embedding or a pretraining objective related to value, such as the outcome of the next action, may be needed before learned embeddings can help.

\subsection{Implications for Evaluating Action Values}
Three aspects of the evaluation changed the conclusions. Comparing the values with the realised returns and with xG showed that the values of the prior setting were close to a constant, which correlations of player-level values with other indicators could not reveal. Requiring the same sign for all seeds removed off-ball differences, in both directions, that a bootstrap over matches alone would have declared significant. Matching the training length suggested that the apparent gain of pose in the action information of the values may partly reflect the decline in the baseline during longer training. Evaluations of action-value models should therefore report the error against realised returns, a reference such as xG for shots, the results of several seeds, and the training length, and should select models on the value error rather than on a loss that includes action prediction or on the TD loss \cite{fujimoto2022should}.

\subsection{Limitations}
The data come from one competition and season, and the test set contains 12 matches. The models were stopped on the total validation loss and the test used the last weights, as in the public implementation of the prior model; without the action loss, all TD runs stopped after six epochs because the minimum improvement of early stopping was larger than the whole decrease of their validation loss after the first epoch, and in a third of these runs the validation loss was still lowest in the last epoch (Section~\ref{sec:training}); in particular, the stability of package TD without the action loss was tested only over these six epochs. We did not test whether longer training, the selection of the weights with the lowest validation error of the values, or a different optimisation of the jointly trained ST-GCN encoder would change the comparison, so the results hold for this training protocol. The pose condition for the qualitative examples was chosen on the test results. The imminent-action analysis used the pose 0.2~s before the action, which may already show the backswing of a kick, so it is short-range anticipation; the pass-success analysis used the pose of the step of the pass, which can include the kick, and only the sequences in which passes and events could be matched one to one (29\% of the training and 40\% of the test sequences), and our xG model uses only basic shot features. The off-ball values were compared only with the realised returns; a step-level comparison with an off-ball indicator such as OBSO, analogous to that with xG for shots, would give them an external reference, provided the indicator is first shown to predict the outcomes in the same data. Only the pose of the target player and a 0.2-second window were used, and no correction for multiple comparisons was applied; such a correction could remove some of the three robust improvements of the action values but not add new ones, so it would not weaken the conclusion that pose rarely improved them. Conversely, the absence of a robust improvement does not show that pose has no effect: the confidence intervals in Fig.~\ref{fig:forest} bound the effects compatible with the data, and an equivalence test with a margin of practical relevance would be needed to conclude that an effect is negligible. Finally, only the values of the observed actions were validated (Section~\ref{sec:problem}); the values of the actions that were not taken, which can fall far outside the range of the returns (Fig.~\ref{fig:cases}) \cite{fujimoto2019off}, were not.

% ======================================================================
\section{Conclusions}
\label{sec:conclusion}
We examined whether adding 3D pose to a SARSA-based MADRL action-valuation model brings the values of the observed actions closer to the realised returns of the attacks, which are an observed goal or conceded goal and otherwise the expected possession value at the end of the attack. Across six representations, three seeds and five learning settings, pose robustly brought these values closer to the returns in only three of 50 comparisons, all with torso and head orientation and in two of the five settings, and learned ST-GCN embeddings often made them worse. The learning setting of the prior model limited the values far more than the state did: its action-prediction loss lowered their validity also at matched training length. The same pose features, in contrast, improved the prediction of the on-ball action 0.2~s ahead, and the pose of the passer at the pass, which can include the kick, was associated with pass success. For our data from one competition and season, the representations and the training protocol examined, with a reward given only at the end of an attack, the improvement of the action values by pose was thus limited. We validated only the values of the observed actions, not the values of the actions that were not taken or the choice of an optimal action. Future work should train action values with outcomes closer in time, such as pass success, ball retention or changes in expected goals, select models on the value error, predict the actions with a separate output head instead of the values themselves, pretrain the pose encoder, include the poses of surrounding players, and verify the results on other competitions.

\appendices

\section{Pose Input and Missing Joints}
\label{app:pose-input}
The 29 joints are the nose, neck and mid-hip, the left and right shoulders, elbows, wrists, hips, knees, ankles, eyes and ears, the big toes, small toes and heels of both feet, and the thumbs and little fingers of both hands. In preprocessing, the coordinates of every joint in a frame are expressed relative to the mid-hip joint of the same frame and divided by 2.44 m on all three axes, so the proportions and angles of the body are kept; a frame without the mid-hip joint is treated as a frame without pose. The pose of a 5 Hz step consists of the at most five 25 Hz frames of the step.

The hand-crafted representations use only the available joints. For the raw representation, each joint is averaged over the frames in which it is present. For the orientation features, the torso orientation of a frame requires both shoulders, and the head orientation is computed only in frames with a torso orientation and requires at least one of the eye and the ear on each side, the left and right points being the mean of the available eye and ear; each feature is averaged over the frames in which it can be computed. A joint or feature that is not available in any frame of the step is set to zero.

The ST-GCN input $P_t$ is a fixed $5\times29\times3$ array without a mask. A missing joint is set to zero, i.e., placed at the mid-hip; a missing frame and a step without pose are set to zero in all joints. The padded steps of a batch are also zero. The encoder is applied to all steps of the batch, including padded steps, and the batch-normalisation statistics are computed over all of them; the losses and metrics use only the real steps.

\section{ST-GCN Encoder}
\label{app:stgcn}
A spatial-temporal graph convolutional network (ST-GCN) \cite{yan2018spatial} processes a sequence of skeletons as a graph whose vertices are the joints of every frame. Spatial edges connect the joints linked by a bone within a frame, and temporal edges connect the same joint in consecutive frames. A spatial graph convolution updates the features of each joint from those of its neighbours in the skeleton, and a temporal convolution then mixes the features of each joint over neighbouring frames; stacking such blocks lets the features cover larger parts of the body and longer intervals, and pooling over the joints and frames gives a fixed-length vector. Because the weights are shared over the joints and frames and the graph follows the skeleton, the network can represent coordinated movements of body parts with few parameters. The original ST-GCN for action recognition stacks nine blocks with up to 256 channels and a temporal kernel of nine frames, partitions the neighbours of each joint into subsets with separate weights, learns weights for the edges, and is trained with a classification loss on labelled movements. Our encoder is a lightweight variant for a window of only five frames: it has two blocks, one normalised adjacency matrix without partitions or learned edge weights, and a temporal kernel of three frames, and it is trained only through the losses of the value model.

The skeleton graph has the 29 joints as vertices and 28 bones as undirected edges: nose--neck, nose--eyes, eyes--ears, neck--mid-hip, neck--shoulders, shoulders--elbows, elbows--wrists, wrists--thumbs and wrists--little fingers, mid-hip--hips, hips--knees, knees--ankles, ankles--heels, and heels--big toes and heels--small toes, on both sides. With $\mathbf{A}$ the adjacency matrix including self-loops and $\mathbf{D}$ its degree matrix, the encoder uses the normalised adjacency matrix $\hat{\mathbf{A}}=\mathbf{D}^{-1/2}\mathbf{A}\mathbf{D}^{-1/2}$ of graph convolutional networks \cite{kipf2017semi}. For an input $\mathbf{X}\in\mathbb{R}^{C\times5\times29}$ (channels, frames, joints), a block computes
\begin{align}
    \mathbf{H} &= \mathrm{ReLU}\bigl(\mathrm{BN}(\mathbf{W}_s\,\mathbf{X}\hat{\mathbf{A}})\bigr), \\
    \mathbf{Y} &= \mathrm{ReLU}\bigl(\mathrm{BN}(\mathrm{Conv}_{3\times1}(\mathbf{H})) + R(\mathbf{X})\bigr),
\end{align}
where $\mathbf{W}_s$ is a $1\times1$ convolution over the channels, $\mathrm{Conv}_{3\times1}$ is a convolution over three consecutive frames with zero padding that keeps the five frames, BN is batch normalisation \cite{ioffe2015batch}, and $R$ is a $1\times1$ convolution with batch normalisation when the number of channels changes and the identity otherwise. The convolutions have no bias, and no dropout is used. The first block maps 3 to 64 channels and the second 64 to 128, and the embedding $g_\phi(P_t)\in\mathbb{R}^{128}$ is the mean of the output over the five frames and the 29 joints. The encoder has 79,360 parameters.

The encoder is initialised with the default initialisation of PyTorch and trained jointly with the value model by the same optimiser, learning rate and weight decay (Section~\ref{sec:method}); it is not pretrained, and no loss acts on it other than the losses of the value model. Table~\ref{tab:params} lists the number of trainable parameters of each condition.

\begin{table}[h]
\centering
\caption{Input dimension and number of trainable parameters of the action-value model for each pose condition (input layer, GRU and output layer with 16 units for the 15 actions and the padding class, plus the ST-GCN encoder where used).}
\label{tab:params}
\begin{tabular}{lrr}
\hline
Condition & Input dim. & Parameters \\
\hline
none & 92 & 31,952 \\
raw & 179 & 37,520 \\
torso & 98 & 32,336 \\
torso+head & 106 & 32,848 \\
ST-GCN & 220 & 119,504 \\
ST-GCN+torso & 226 & 119,888 \\
\hline
\end{tabular}
\end{table}

\bibliographystyle{IEEEtran}
\bibliography{refs}

\begin{IEEEbiography}
[{\includegraphics[width=1in,height=1.25in,clip,keepaspectratio]{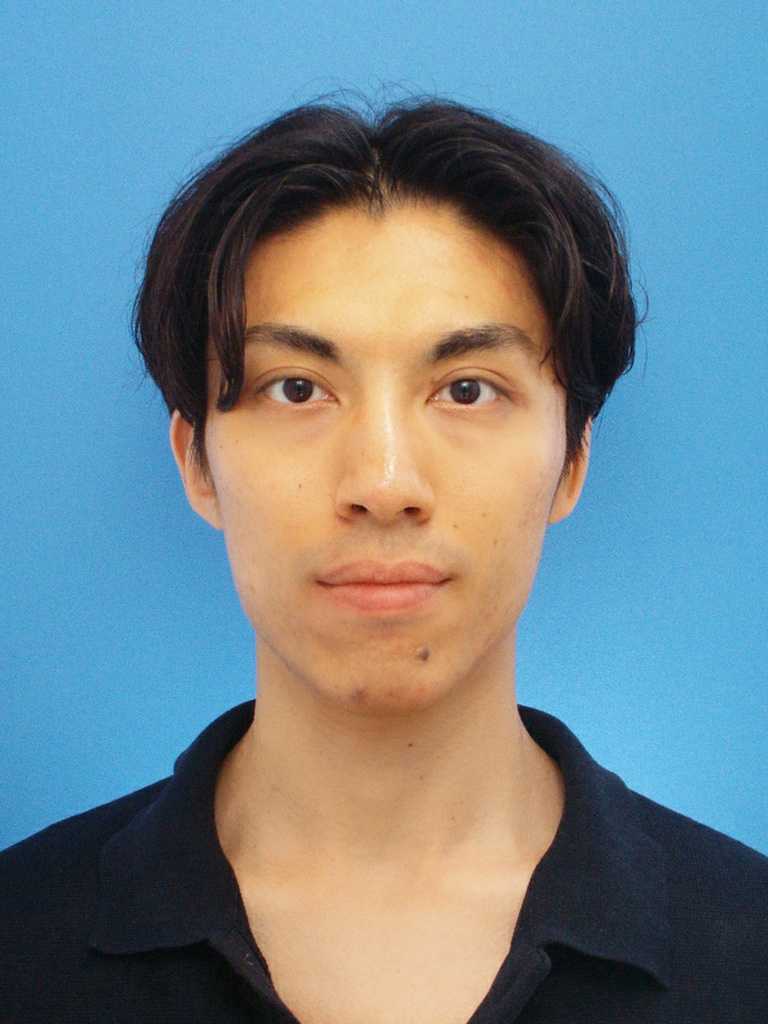}}]{Rikuhei Umemoto} received his B.E. degree from Kyoto University, Kyoto, Japan, in 2022. He also received his M.S. degree from Nagoya University, Nagoya, Japan. 
His research interests include data analysis with computer vision and reinforcement learning in team sports such as football.
\end{IEEEbiography}

\begin{IEEEbiography}[{\includegraphics[width=1in,height=1.25in, clip,keepaspectratio]{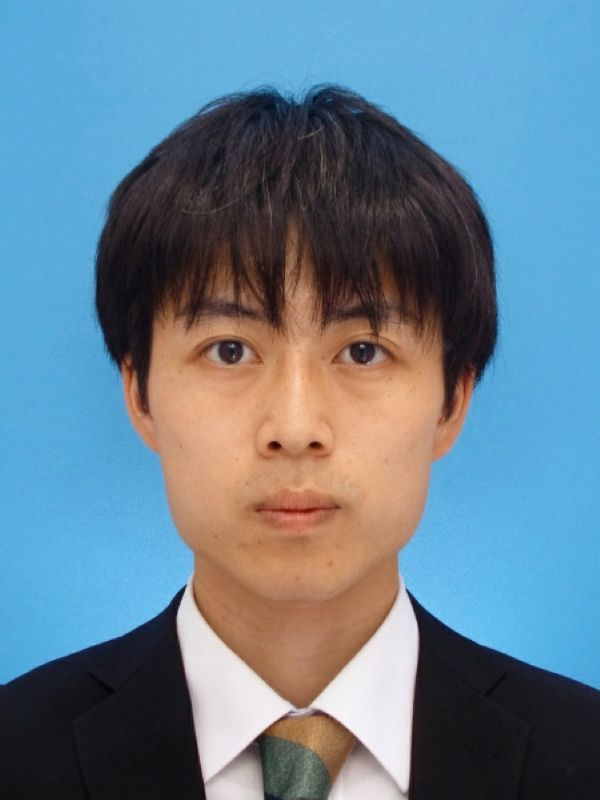}}]{Kenjiro Ide} received his B.S. and M.S. degrees from Nagoya University in 2024 and 2026, respectively.
His research interests are multi-agent reinforcement learning and generative modeling for understanding and evaluating decision-making in team sports, particularly football.
\end{IEEEbiography}

\begin{IEEEbiography}[{\includegraphics[width=1in,height=1.25in, clip,keepaspectratio]{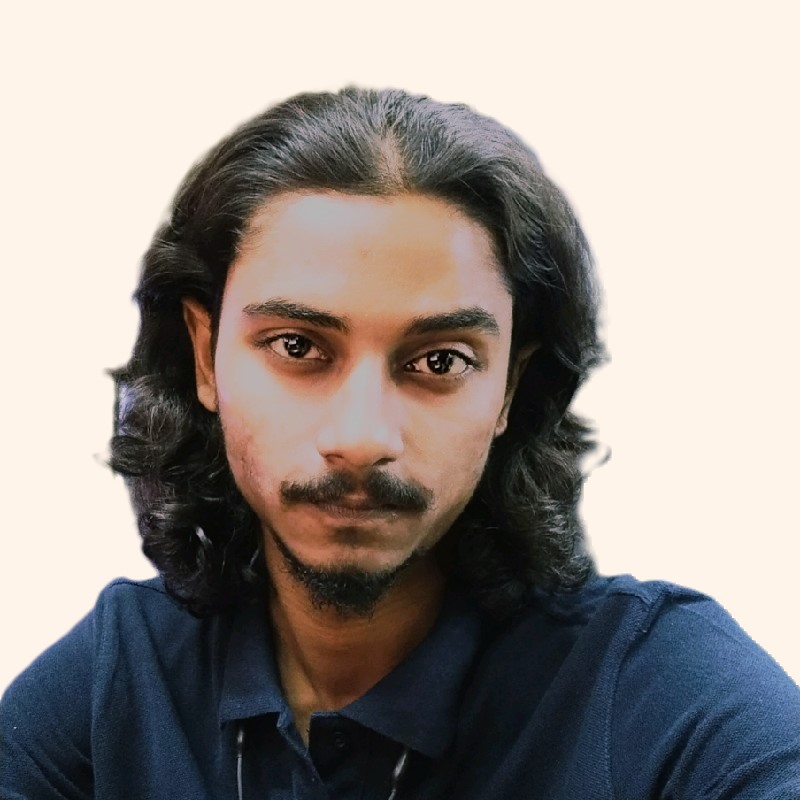}}]{Soham Majumder} received his B.E. degree from St. Xavier's College, Kolkata, India, in 2022. He also received his M.S. degree from the Indian Statistical Institute, India, in 2023.
His research interests include spatiotemporal data analysis, explainable machine learning, graph neural networks, and their applications to football analytics.
\end{IEEEbiography}

\begin{IEEEbiography}[{\includegraphics[width=1in,height=1.25in, clip,keepaspectratio]{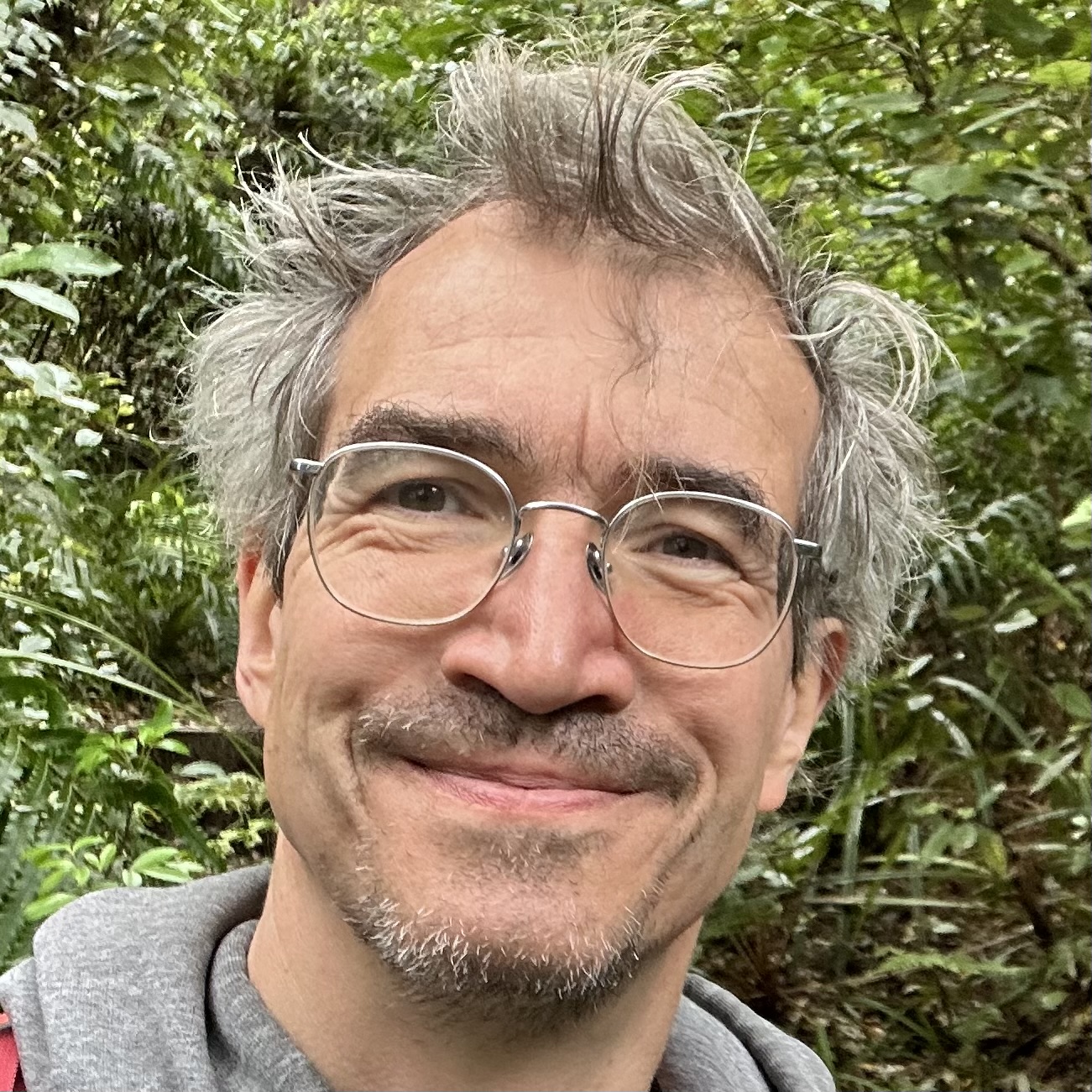}}]{Ulf Brefeld} received his Diploma degree in Computer Science from Technische Universität Berlin, Germany, in 2003 and his Ph.D. degree from Humboldt-Universität zu Berlin, Germany, in 2008. He previously worked on machine learning at several academic and industrial research institutions, including Universität Bonn, Yahoo! Research Barcelona, and the Max Planck Institute for Computer Science. He also led the Recommender Systems group at Zalando SE and served as a joint professor of Knowledge Mining and Assessment at TU Darmstadt and the Leibniz Institute for Research and Information in Education (DIPF). He is currently a professor of Machine Learning at Leuphana Universität Lüneburg, Germany. 
His research interests are statistical machine learning, data mining, spatiotemporal data analysis, recommender systems, and sports analytics.
\end{IEEEbiography}

\begin{IEEEbiography}[{\includegraphics[width=1in,height=1.25in, clip,keepaspectratio]{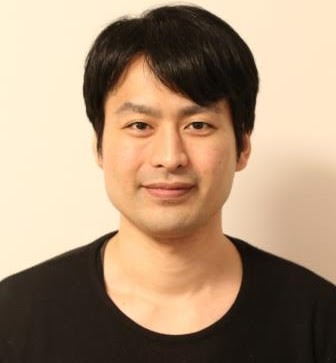}}]{Keisuke Fujii} received his B.S., M.S., and Ph.D. degrees from Kyoto University in 2009, 2011, and 2014, respectively.
After his work as a postdoctoral fellow and research scientist at Nagoya University and RIKEN Center for Advanced Intelligence Project in Japan, he joined Nagoya University.
He is currently a professor at the Information Technology Center and Graduate School of Informatics.
His research interests are interdisciplinary studies among machine learning, behavioral sciences, and sports sciences.
\end{IEEEbiography}

\EOD

\end{document}